\documentclass{article}

\usepackage{amsmath,amsfonts,bm}

\def\eqref#1{equation~\ref{#1}}

\def\1{\bm{1}}

\DeclareMathAlphabet{\mathsfit}{\encodingdefault}{\sfdefault}{m}{sl}
\SetMathAlphabet{\mathsfit}{bold}{\encodingdefault}{\sfdefault}{bx}{n}

\usepackage{microtype}
\usepackage{graphicx}
\usepackage{subcaption}
\usepackage{caption}
\usepackage{booktabs} 
\usepackage{hyperref}
\usepackage[capitalize,noabbrev]{cleveref}
\usepackage{duckuments}
\usepackage{multirow}
\usepackage{makecell}
\usepackage{enumitem}
\usepackage{colortbl}
\usepackage{xspace}
\usepackage{xcolor}
\usepackage{lipsum}
\usepackage{amssymb}
\usepackage{mathtools}
\usepackage{amsthm}
\usepackage{bbm}
\usepackage{listings}
\usepackage{wrapfig}
\usepackage{float}
\usepackage{minted}

\definecolor{RowHighlight}{gray}{0.9}

\usepackage[preprint]{icml2026}

\theoremstyle{plain}

\theoremstyle{definition}

\theoremstyle{remark}

\usepackage[textsize=tiny]{todonotes}

\newcommand{\llname}{Beyond Correctness: Learning Robust Reasoning via Transfer}
\newcommand{\lname}{Beyond Correctness: Learning Robust Reasoning via Transfer}
\newcommand{\mname}{Reinforcement Learning with Transferable Reward (RLTR)\xspace}
\newcommand{\mmname}{Reinforcement Learning with Transferable Reward}
\newcommand{\sname}{RLTR\xspace}

\newcommand{\placeholder}[1]{{\color{lightgray}\lipsum[1]}}
\newcommand{\icmlInternship}{\textsuperscript{$\dagger$} Work done during an internship at Microsoft Research Asia}
\icmltitlerunning{\lname}

\begin{document}

\twocolumn[
    \icmltitle{\llname}
    
    \icmlsetsymbol{equal}{*}
    \icmlsetsymbol{intern}{$\dagger$}
    
    \begin{icmlauthorlist}
    \icmlauthor{Hyunseok Lee}{kaist,intern}
    \icmlauthor{Soheil Abbasloo}{msr}
    \icmlauthor{Jihoon Tack}{msr}
    \icmlauthor{Jinwoo Shin}{kaist}
    \end{icmlauthorlist}
    
    \icmlaffiliation{kaist}{KAIST}
    \icmlaffiliation{msr}{Microsoft Research}

    \icmlkeywords{LLM, LLM reasoning, Test-time scaling}

    \icmlcorrespondingauthor{Soheil Abbasloo}{soheil.abbasloo@microsoft.com}
    
    \vskip 0.3in
]

\printAffiliationsAndNotice{\icmlInternship} 
\begin{abstract}

Reinforcement Learning with Verifiable Rewards (RLVR) has recently strengthened LLM reasoning, but its focus on final‑answer correctness leaves a critical gap: it does not ensure the robustness of the reasoning process itself. 
We adopt a simple philosophical view—robust reasoning should remain useful beyond the mind that produced it—and treat reasoning as a form of meaning transfer that must survive truncation, reinterpretation, and continuation. 
Building on this principle, we introduce \mname, which operationalizes robustness via transfer reward that tests whether a partial reasoning prefix from one model can guide a separate model to the correct answer. 
This encourages LLMs to produce reasoning that is stable, interpretable, and genuinely generalizable. 
Our approach improves sampling consistency while improving final‑answer accuracy, and it reaches comparable performance in substantially fewer training steps. 
For example, on MATH-500, \sname achieves a +3.6\%p gain in Maj@64 compared to RLVR and matches RLVR’s average accuracy with roughly $2.5\times$ fewer training steps, providing both more reliable reasoning and significantly more sample-efficient.

\end{abstract}
\section{Introduction}
\label{sec:intro}

Large language models (LLMs) combined with reinforcement learning (RL) have recently demonstrated remarkable performance across various domains, such as solving complex mathematical problems~\citep{shao2024deepseekmath}, 
serving as coding assistants~\citep{le2022coderl}, and contributing to scientific discoveries~\citep{romera2024discovery}.
Compared with simple supervised fine-tuning, RL enables LLMs to learn more generalizable and robust behaviors~\citep{ouyang2022instructgpt}.
However, the success of RL in LLMs is often bottlenecked by its heavy reliance on reward models.
While reward models reduce the need for costly human annotations, they introduce critical challenges such as reward hacking~\citep{skalse2022rewardhack} and increased training complexity~\citep{casper2023openproblem}.

To tackle this issue, reinforcement learning with verifiable reward (RLVR) has recently emerged~\citep{shao2024deepseekmath,guo2025deepseekr1}.
RLVR replaces learned reward models with rule-based verification of the correctness of final answers, which significantly simplifies the training pipeline and alleviates reward hacking and the need for extensive human annotations. 
As a result, RLVR achieves remarkable success on complex reasoning tasks, especially in competitive math domains.
Yet despite these gains, RLVR optimizes reasoning solely through final‑answer correctness, leaving a critical dimension of reasoning quality underexplored: the robustness of the reasoning process itself. 
In real‑world and practical settings, models that consistently produce correct answers across different sampled solutions, decoding variations, or reasoning paths are far more valuable than models that succeed only sporadically. 
However, RLVR’s objective evaluates only the generator’s final output, providing no incentive for intermediate reasoning to be robust, reusable, or stable under such perturbations. 
Prior work has observed that the model loses consistency and diversity as more samples are generated, suggesting that final-outcome correctness alone may be an incomplete signal for promoting reliable reasoning behavior~\citep{yue2025does}.

\begin{figure*}[t]
  \includegraphics[width=\textwidth]{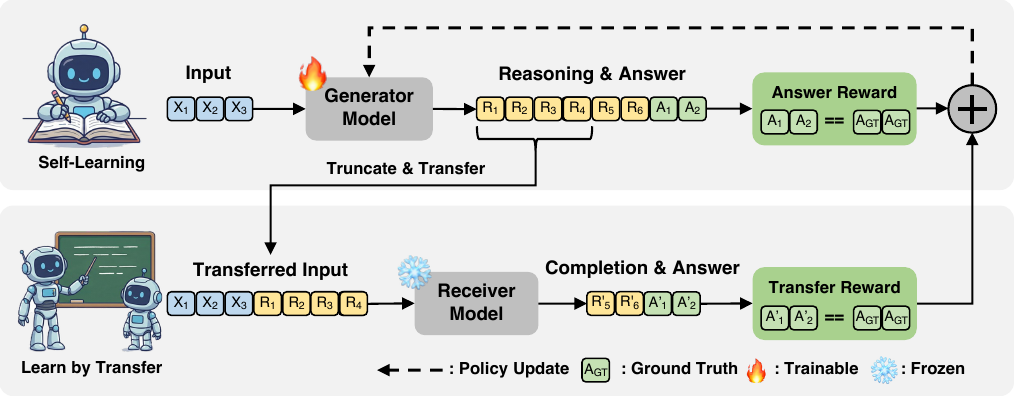}
  \caption{ \textbf{Overview of \sname: }
  \sname augments standard RLVR with a transfer reward. \textbf{Top:} A trainable generator model produces a full completion, whose final-answer correctness yields on answer reward for policy optimization. \textbf{Bottom:} We then truncate the generated reasoning to form a prefix and feed it to a frozen receiver model to produce continued completions whose final-answer correctness defines a transfer reward that measures the transferability of partial reasoning across models. The answer reward and transfer reward are combined into a unified reward signal used to update the generator policy.
  } \label{figure:method_overview}
\end{figure*}

To move beyond outcome‑only evaluation, we draw inspiration from human learning: strong reasoning is not merely correct—it is also \emph{explainable} in a way that others can reliably follow.  
This motivates our central concept, \emph{reasoning transferability}, defined as the extent to which a reasoning prefix produced by one model can be completed by another model to reach a verifiably correct answer. 
We operationalize this idea by truncating a generator’s reasoning trace and asking a separate receiver model to continue it; transfer succeeds when the combined reasoning yields a correct final answer.  
Building on this principle, we propose \mname, a new reinforcement learning framework that explicitly optimizes cross‑model reasoning transferability.  
The core mechanism is a \emph{transfer reward} that evaluates a generator’s partial reasoning via a receiver‑model continuation, integrated into RLVR through a transfer roll‑out procedure and a weighted combination of answer, transfer, and format rewards.  
To our knowledge, \sname is the first RLVR framework that directly incentivizes LLMs to produce reasoning traces that are transferable across models. 
An overview of \sname~is shown in Figure~\ref{figure:method_overview}.

We demonstrated the effectiveness of \sname through extensive evaluations on multiple reasoning datasets covering a wide range of difficulties across mathematical and scientific domains. 
Overall, \sname consistently improves Majority Voting at K (Maj@K), a proxy for multi-sample consistency, while also improving average accuracy.
For instance, on AMC23~\citep{maa23amc}, \sname improves Maj@64 from 61.7 to 67.5 compared to RLVR.
Across training, \sname achieves the same average accuracy level with about $2.5\times$ fewer training steps than RLVR on MATH-500, indicating faster learning and improved sample efficiency.
Finally, we show transferability correlates with multi-sample consistency by tracking transferability and Maj@$K$ throughout training.

We highlight the main contributions of the paper below:
\begin{itemize}[itemsep=0.01in,leftmargin=0.2in,topsep=0.02in, nosep]
    \item{We introduce \sname, a simple extension of RLVR that adds a transfer reward to optimize reasoning transferability across models.}
    \item{We demonstrate that \sname increases consistency, as shown by higher majority voting accuracy, which reflects more robust reasoning compared to RLVR.}
    \item{Our training framework improves training sample efficiency e.g., \sname matches RLVR's performance with $2.5\times$ fewer training steps and improved average accuracy on MATH-500.}
\end{itemize}
\section{Related Work}
\label{sec:related}

\textbf{Reinforcement Learning in LLM.}
Reinforcement learning (RL) has become a cornerstone for aligning LLMs with human intent, primarily through Reinforcement Learning from Human Feedback (RLHF; \citealp{ouyang2022instructgpt}). 
Standard approaches typically employ Proximal Policy Optimization (PPO; \citealp{schulman2017ppo}) to optimize a policy against a learned reward model (RM) that mimics human preferences. 
However, this dependence on neural reward models introduces significant challenges. 
As the policy optimizer exploits the reward model, it often leads to reward hacking, where the model generates high-reward but low-quality or nonsensical outputs due to the imperfections of the proxy reward \citep{skalse2022rewardhack,gao2023scalinghack,saito2023verbosity}. 
Furthermore, training a robust reward model requires extensive human annotation and often creates a bottleneck in scaling reasoning capabilities.

\textbf{Reinforcement Learning with Verifiable Reward.} 
To mitigate the limitations of neural reward models, recent research has shifted towards RL with Verifiable Reward (RLVR), where the reward is derived from ground-truth correctness (e.g., mathematical answers or code execution) rather than a learned proxy. 
Pioneering works such as DeepSeek-Math \citep{shao2024deepseekmath} and DeepSeek-R1 \citep{guo2025deepseekr1} introduced Group Relative Policy Optimization (GRPO) to stabilize RL training without a critic model, with verifiable reward that can induce emergent reasoning behaviors, significantly boosting performance on complex benchmarks. 
The following works were mainly focusing on stabilize the learning algorithm and enhance the learning objectives \citep{yu2025dapo, liu2025drgrpo}.
While these methods effectively address reward hacking by relying on objective outcomes, they primarily evaluate the result rather than the process. 
Unlike prior approaches that rely solely on self-consistency or final-answer correctness, \sname introduces a novel reward signal based on \textit{reasoning transferability}. 

\textbf{Process Reward Models.}
Process Reward Models (PRMs) offer an alternative by providing dense step-level supervision using
learned verifiers trained on large annotated reasoning traces.
Process supervision has been shown to reduce reasoning errors compared to 
outcome-only training~\citep{uesato2022processOutcome}.
Verifier-based step scoring has been explored for GSM8K~\citep{cobbe2021gsm8k}, 
and large-scale datasets (e.g., PRM800K) enable powerful verifier models 
that outperform outcome-only training on MATH~\citep{lightman2023lets}.
Recent variants include Q-value PRMs~\citep{li2024pqm}, 
reasoning-driven PRMs~\citep{she2025rprm}, and generative verifier models 
for long chain-of-thought~\citep{khalifa2025thinkprm}.  
However, PRMs require high-capacity learned step-level verifiers, substantial 
step-level annotations, and reintroduce proxy reward mismatch.
Unlike PRMs, \sname operates without any additional annotation requirements beyond those of the answer-verification dataset, and does not require training any reward models for intermediate reasoning supervision.

\section{\sname: \mmname} \label{sec:method}

In this section, we introduce \mname, a reinforcement learning framework that encourages robust and reusable reasoning traces by augmenting RLVR with a transfer reward. 
Building on RLVR-style final answer supervision, \sname introduces a transfer reward that evaluates whether another model (namely, the receiver model) can recover the correct answer from a truncated reasoning prefix, 
and maximizes this signal as an auxiliary reward objective.
We first formalize the learning framework, RLVR, in Section~\ref{sec:method_prelim} and at the top of Figure~\ref{figure:method_overview}.
We then present \sname, the core training algorithm that integrates the transfer reward into the RL objective (Section~\ref{sec:training} and bottom in Figure~\ref{figure:method_overview}).

\newcommand{\model}{\mathcal{M}}
\newcommand{\gmodel}{\model_\text{gen}}
\newcommand{\rmodel}{\model_\text{rcv}}
\newcommand{\gt}{y_\text{gt}}
\newcommand{\areward}{\text{R}_{\text{ans}}}
\newcommand{\treward}{\text{R}_{\text{trans}}}
\newcommand{\freward}{\text{R}_{\text{fmt}}}

\subsection{Preliminary: RLVR}
\label{sec:method_prelim}

RLVR optimizes the model to produce a correct final answer under a fixed output format.
Given an input $x$, the generator model $\gmodel$ samples a full response (including reasoning and the final answer), i.e., $y_{\text{gen}}\sim\gmodel(\cdot|x)$.
The reward is primarily defined by an answer reward that checks whether the generator's final answer matches the ground truth, i.e.,
$\areward(y_{\text{gen}})=\mathbbm{1}[\mathrm{answer}(y_{\text{gen}})=\gt]$.
In addition, a format reward $\freward(y_{\text{gen}})$ is used to encourage well-formed outputs (e.g., valid reasoning structure and answer formatting), and to penalize invalid generations.

The overall reward is defined as
\begin{equation}\label{eq:rlvr_reward}
\text{R}_{\text{RLVR}}(y_{\text{gen}})= a\,\areward(y_{\text{gen}}) + \freward(y_{\text{gen}}).
\end{equation}

For a given dataset $\mathcal{D}:=\{(x,y)\}$, the generator is optimized via reinforcement learning with the objective as:
\begin{equation}
\max_{\gmodel}\; 
\mathbb{E}_{(x,y_{\text{gt}})\sim \mathcal{D}}\left[\mathbb{E}_{y_{\text{gen}}\sim\gmodel(\cdot|x)}
\left[ \text{R}_{\text{RLVR}}(y_{\text{gen}}) \right]\right].
\end{equation}

In this paper, we mainly optimize the RL policy with group relative policy optimization (GRPO; \citealp{shao2024deepseekmath, guo2025deepseekr1}), which is known to be simple yet strong when use with RLVR. 

Overall, conventional RLVR optimizes the correctness of the generator's own final output. 
However, it does not explicitly account for step-level noise and brittleness in the reasoning trajectory, nor does it directly encourage reusable or interpretable reasoning that can be reliably continued by other models or decoding processes.

\subsection{\mmname}\label{sec:training}

To address this limitation, we propose \sname, a training method that augments RLVR with a transfer reward to encourage reasoning \emph{transferable}, whether partial reasoning that can be reliably reused by another model to reach the correct answer.
Intuitively, a generator has higher transferability when its truncated reasoning prefixes contain stable and informative intermediate structure, enabling a receiver to reliably complete the solution, rather than relying on the generator's own continuation.

\textbf{Transfer Reward.}
We define transfer reward which identify the reasoning transferability.
Concretely, given an input $x$, the generator model $\gmodel$ produces a full reasoning sequence $y_{\text{gen}}\sim\gmodel(\cdot|x)$. Then, we truncate the prefix of reasoning sequence and passed to a receiver model $\rmodel$ for continuation, $y_{\text{rcv}}\sim\rmodel(\cdot|x,y_{\text{gen}}[:\ell])$, where truncation ratio $\tau \in (0,1]$ and truncation length $\ell = \lfloor \tau \, |y_{\text{gen}}| \rfloor$.
We define a transfer reward as:
\begin{equation}\label{eq:transfer_reward}
\treward(y_{\text{gen}},\rmodel)=\mathbbm{1}[\mathrm{answer}(y_{\text{rcv}})=\gt]
\end{equation}
which indicates whether the receiver can reach the correct answer from the truncated reasoning.
In practice, we compute $\treward$ using a single continuation, and sample $\tau$ uniformly between 0.3 and 0.9 to ensure the model learns reasoning pattern robust arbitrary interruptions, rather than overfitting to fixed truncation boundaries.

The overall reward is defined as
\begin{equation}\label{eq:rltr_reward}
\text{R}_{\text{RLTR}}(y_{\text{gen}},\rmodel):=\text{R}_{\text{RLVR}}(y_{\text{gen}}) + t\,\treward(y_{\text{gen}},\rmodel)
\end{equation}

\textbf{Training Objective.} The generator is optimized via reinforcement learning with the objective
\begin{equation}
\max_{\gmodel}\; 
\mathbb{E}_{(x,y_{\text{gt}})\sim \mathcal{D}}\left[\mathbb{E}_{y_{\text{gen}}\sim\gmodel(\cdot|x)}\left[ \text{R}_{\text{RLTR}}(y_{\text{gen}},\rmodel) \right]\right].
\end{equation}

\section{Experiments}
\label{sec:experiments}

\begin{figure*}[t]
  \centering
    \begin{subfigure}[t]{0.33\textwidth}
    \centering
    \includegraphics[width=\linewidth]{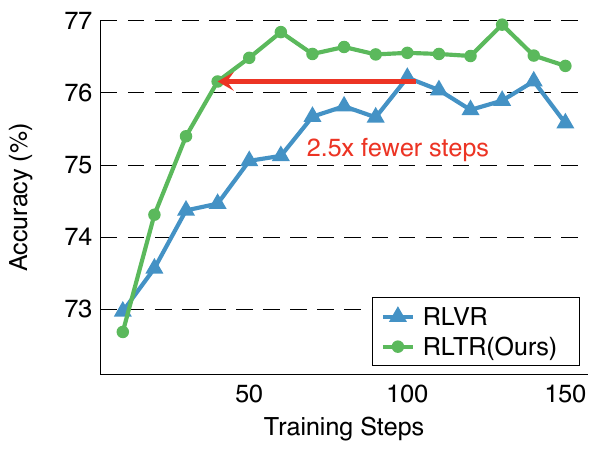}
    \caption{\textbf{Average accuracy}}\label{fig:average_progress}
    \label{fig:avg_progress}
  \end{subfigure}\hfill
  \begin{subfigure}[t]{0.33\textwidth}
    \centering
    \includegraphics[width=\linewidth]{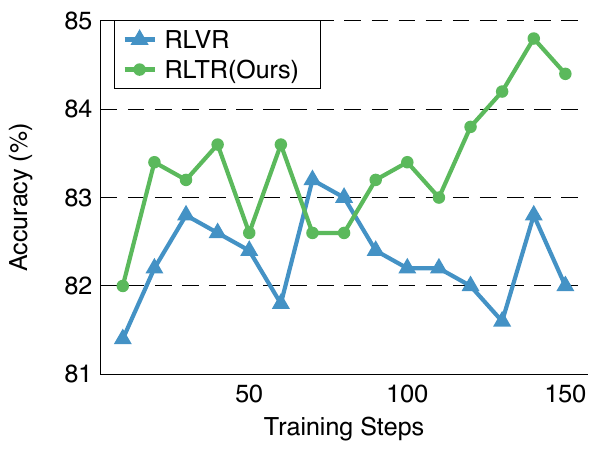}
    \caption{\textbf{Maj@64 accuracy}}
    \label{fig:maj_progress}
  \end{subfigure}\hfill
  \begin{subfigure}[t]{0.33\textwidth}
    \centering
    \includegraphics[width=\linewidth]{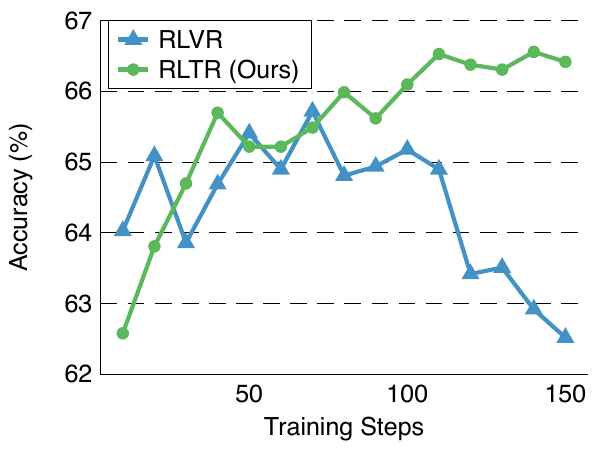}
    \caption{\textbf{Transferability}}
    \label{fig:transfer_progress}
  \end{subfigure}
  \caption{
  \textbf{Training dynamics and transferability.}
    \textbf{(a)} \textbf{Average accuracy}: Our method \sname, significantly reduce the training steps to achieve comparable performance with RLVR-trained models. Notably, \sname matches RLVR with $2.5\times$ fewer training steps.
  \textbf{(b)} \textbf{Maj@64}: Compared to RLVR, \sname progressively improves consistency over training, whereas RLVR’s performance degrades as training proceeds.
  \textbf{(c)} \textbf{Transferability}: 
  Transferability follows Maj@64, supporting our view that optimizing the transfer reward through \sname improves reasoning consistency and robustness.
  Across all three views, \sname consistently outperforms \textsc{RLVR}, indicating a shift toward reliably correct dominant solution modes and more stable optimization progress under the proposed transfer-based reward.
  }
  \label{fig:training_and_transfer}
  \vspace{-0.07in}
\end{figure*}

We provide an empirical evaluation of Transferability and \sname by investigating the following questions:
\begin{itemize}[leftmargin=*,topsep=0.0pt,itemsep=.5pt]
    \item \textbf{RQ1}: Can \sname improve accuracy and consistency? (Table~\ref{tab:main_table}, Table~\ref{tab:ablation_llama}, and Table~\ref{tab:ood_gpqa})
    \item \textbf{RQ2}: How does \sname affect training dynamics and efficiency? (Figure~\ref{fig:training_and_transfer} and Table~\ref{tab:cost_analysis})
    \item \textbf{RQ3}: Is transferability correlated with consistency of reasoning process? (Figure~\ref{fig:training_and_transfer})
    \item \textbf{RQ4}: How does the proposed components contribute to performance. (Table~\ref{tab:reward_ratio} and Table~\ref{tab:receiver_ablation})
\end{itemize}

\textbf{Evaluation setup.} We mainly report majority voting at K (Maj@K) and average accuracy (Acc.) as a sampling-based metric. 
Maj@K selects the final answer by taking the most frequent answer among the K samples, and counts it as correct if this majority answer matches the ground truth. 
Specifically, Maj@K captures reasoning consistency and robustness by evaluating whether the correct answer dominates across K stochastic generations.
Average accuracy is expectation of the single-sample's accuracy, which acquired by simply averaging the accuracy of K sampled solutions.
Additionally, in analysis we measure transferability as the average accuracy under a cross-model reasoning transfer continuation.
For each model and benchmark, we repeat sampling-based evaluation three times with different random seeds and report the average.
We evaluate our method on MATH-500~\citep{hendrycks2021math}, GSM8K~\citep{cobbe2021gsm8k}, AMC23~\citep{maa23amc}, and AIME2024~\citep{maa24aime}, which are widely used mathematical reasoning benchmarks, and additionally include the challenging scientific reasoning benchmark GPQA~\citep{rein2024gpqa}.
During the evaluation, we generate with consistent sampling hyperparameter, temperature  $T=1.0$. 
We report the specified evaluation setup details in Appendix~\ref{app:evalution_detail}.

\begin{table*}[tbp] 
    \caption{\textbf{Performance of \sname.} Average accuracy (\%) and Maj@K (\%) of \sname (Ours) compared with baselines, including the base model and RLVR-trained model. We evaluate on two medium-difficulty math reasoning benchmarks, MATH-500~\citep{hendrycks2021math} and GSM8K~\citep{cobbe2021gsm8k}, as well as two more challenging benchmarks, AIME2024~\citep{maa24aime} and AMC23~\citep{maa23amc}. We report average accuracy (Acc.) over 64 samples and majority@K (Maj@K), where Maj@K selects the final answer by majority voting among $K$ samples. Bold values denote the best performance within each group. \sname consistently outperforms other baselines. \textbf{Notably, the performance gains of \sname are substantially larger on the harder and more challenging benchmarks.}} 
    \label{tab:main_table}
    \begin{subtable}{\textwidth}
    \caption{Moderate-difficulty tasks}\label{tab:main_moderate}
    \vspace{-.03in}
    \small
    \centering
    \label{tab:main}
        \begin{tabular}{l cccc cccc}
        \toprule
        & \multicolumn{4}{c}{MATH-500\scriptsize{\textit{ (In-distribution)}}} & \multicolumn{4}{c}{GSM8K\scriptsize{\textit{ (Out-of-distribution)}}} \\
        \cmidrule(lr){2-5}\cmidrule(lr){6-9}

        Metrics&Acc.& Maj@4& Maj@16& Maj@64&Acc.& Maj@4& Maj@16& Maj@64\\
        \midrule
        Base model&71.0& 78.0& 81.2& 82.6&89.1&91.5&  92.7& 93.3\\ 
        RLVR&76.2& 78.2& 80.2& 82.2&
        89.4&90.9& 92.5&92.9\\ 
        \rowcolor{RowHighlight}\textbf{\sname (Ours)}& \textbf{77.0}& \textbf{79.0}& \textbf{83.8}& \textbf{84.2}& \textbf{92.0}& \textbf{93.4}& \textbf{94.0}& \textbf{94.2}\\
        \bottomrule
        \end{tabular}
    \vspace{.07in}
    \end{subtable}
    \begin{subtable}{\textwidth}
    \caption{Hard and challenging tasks}
    \vspace{-.03in}
    \label{tab:ood_aime}
    \small
    \centering
        \begin{tabular}{l cccc cccc}
        \toprule
        & \multicolumn{4}{c}{AIME2024\scriptsize{\textit{ (Out-of-distribution)}}} & \multicolumn{4}{c}{AMC23\scriptsize{\textit{ (Out-of-distribution)}}} \\
        \cmidrule(lr){2-5}\cmidrule(lr){6-9}
        Metrics&Acc.& Maj@4& Maj@16& Maj@64&Acc.& Maj@4& Maj@16& Maj@64\\
        \midrule
        Base model&\phantom{0}9.8 & 10.0& 13.3& 16.7&46.2&51.7&  59.2& 60.8\\ 
        RLVR&11.6 & 14.4& 17.8& 18.9&52.8&55.8& 62.5&61.7\\ 
        \rowcolor{RowHighlight}\textbf{\sname (Ours)}& \textbf{14.8} & \textbf{18.9} & \textbf{20.0}& \textbf{21.1}& \textbf{53.5}& \textbf{59.2}& \textbf{66.7}& \textbf{67.5}\\
        \bottomrule
        \end{tabular}
    \end{subtable}
    \vspace{-.07in}
\end{table*}

\textbf{Training setup.} For the main experiment, we trained \sname on Qwen2.5-7B-Instruct~\citep{Yang2024Qwen25TR} checkpoint, which is provided in HuggingFace. Unless stated otherwise, we use Qwen2.5-3B-Instruct~\citep{Yang2024Qwen25TR} as the receiver model $\rmodel$. For analysis, we additionally consider alternative receivers (e.g., Qwen2.5-1.5B-Instruct~\citep{Yang2024Qwen25TR} and Llama3.2-3B-Instruct~\citep{grattafiori2024llama3}). We perform RL training on a 3K subset of the MATH~\citep{hendrycks2021math} training set, restricted to problems with difficulty $\geq 3$ following~\citet{zeng2025simplerl}. 
For training \sname, we adopt reward weight $a=0.1$ and $t=1.0$ from Eq.~\ref{eq:rlvr_reward} and Eq.~\ref{eq:rltr_reward}. 
Additional implementation details are provided in Appendix~\ref{app:training_detail}.

\textbf{Baseline.} We compare \sname the base model (the initial checkpoint before RL) and RLVR, which applies conventional reinforcement learning with verifiable reward by using the same generator initialization, training data, and decoding/evaluation protocol with \sname. We mostly followed the baseline training setup from~\citet{zeng2025simplerl}.

\subsection{Performance of \sname}
\label{sec:experiment_main}

We present the main result by comparing the performance on math problem-solving. 
Here, we mainly evaluate \sname against RLVR and the base model, reporting average accuracy (Acc.) and Majority at K (Maj@K) to assess performance and consistency.
The main results were evaluated in (i) two moderate-difficulty tasks: MATH-500 and GSM8K and (ii) two challenging tasks: AIME2024 and AMC23.

\textbf{\sname performance gain on moderate-difficulty tasks.} 
We first evaluate \sname on two moderate-difficulty tasks that one is in-distribution benchmark (i.e., MATH-500~\citep{hendrycks2021math}) and the other is out-of-distribution benchmark (i.e., GSM8K~\citep{cobbe2021gsm8k}).
As shown in Table~\ref{tab:main_moderate}, overall, \sname consistently outperforms RLVR across all benchmarks and metrics (e.g., at average accuracy and Maj@K).
It is worth noting that while RLVR achieves strong single-sample average accuracy performance (71.0$\to$76.2 on MATH-500), its multi-sample consistency become less pronounced than those of the base model in Maj@K at larger K, which is consistent with findings in prior work~\citep{yue2025does}.
For instance, RLVR underperforms the base model at higher K: Maj@16 decreases from 81.2 (base) to 80.2 (RLVR).
This suggests that RLVR can produce reasoning trajectories that are brittle. As more samples are drawn, variations in the reasoning paths leads to conflicting outcomes rather than converging toward a consistent correct answer, which in turn manifests as reduced robustness under majority‑vote evaluation.
In contrast, \sname yields substantially better high-K multi-sample aggregation behavior, mitigating RLVR's degradation in Maj@K at large K and consistently outperforming both RLVR and the base model.
For example, on MATH-500, \sname improves Maj@16 from 81.2 to 83.8 and Maj@64 from 82.6 to 84.2, and also improves average accuracy from 71.0 to 77.0.
The stronger gains in Maj@K suggesting that the added transfer reward provides a direct learning signal for robust intermediate reasoning rather than only final-answer correctness.

We observe the same, and even clearer, trend on the out-of-distribution math benchmark, GSM8K. 
As shown in Table~\ref{tab:main}, while RLVR slightly improves single-sample average accuracy, its majority-voting performance degrades at larger K compared to the base model.
In contrast, \sname improves both accuracy and consistency across all metrics (e.g., Acc.: 89.1$\to$92.0), and consistently outperforms RLVR for all K.
This consistent improvements on an out-of-distribution benchmark suggests that the transfer reward provides a general process-level supervision signal, improving robustness beyond the training distribution rather than overfitting to training dataset specific patterns.

\textbf{\sname performance gain on harder tasks.} 
We further evaluate \sname on two challenging math reasoning benchmarks that are out-of-distribution: AIME2024~\citep{maa24aime} and AMC23~\citep{maa23amc}.
AIME2024 and AMC23 are competition-level math benchmarks that are substantially harder than MATH-500~\citep{hendrycks2021math} and GSM8K~\citep{cobbe2021gsm8k}.
As shown in Tables~\ref{tab:ood_aime}, \sname consistently outperforms both the base model and RLVR across sampling-based metrics, indicating that the gains are not limited to the moderate-difficulty math domain benchmarks.
Notably, the improvements are more pronounced on the harder math benchmarks.
On AIME2024, \sname boosts average accuracy from 9.8 to 14.8 and improves Maj@64 from 16.7 to 21.1, whereas RLVR provides little to low gain at high K (half of gain at Maj@64).
A similar trend holds on AMC23, while RLVR yields modest gains, \sname achieves a larger improvement in majority voting, suggesting more reliable aggregation under challenging settings.
Overall, these results suggest that optimizing the transfer reward encourages more robust and reusable reasoning trajectories that generalize better under distribution shift, with particularly strong benefits on harder reasoning tasks.
We believe that this effect is amplified on competition-level problems, where correctness depends more critically on maintaining a coherent reasoning process.
In this settings, the transfer-based signal provides useful process-level supervision beyond final-answer correctness, leading to large gains on harder reasoning tasks.

\subsection{Training Dynamics of \sname}

To better understand the effect of the proposed transfer reward, we analyze the training dynamics of \sname using the following three metrics: (i) average accuracy, (ii) Maj@64 accuracy, and (iii) reasoning transferability.
Here, we evaluate intermediate checkpoints on MATH-500 and compare \sname with RLVR.

\textbf{\sname improves sample efficiency.}
As shown in Figures~\ref{fig:avg_progress}, \sname improves substantially faster than RLVR in average accuracy, and also attains a higher final accuracy, consistent with the overall results in Table~\ref{tab:main}. For example, \sname require about $2.5\times$ fewer training steps to reach a comparable accuracy with RLVR.
This result suggest that the transfer reward provides a richer process-level signal that incentivizes robust reasoning trajectories, helping the model converge faster than RLVR.
Moreover, as shown in Figure~\ref{fig:maj_progress}, the Maj@64 curve also follows this improvement, indicating that the gains are not driven by sporadic correct samples but by a progressively more reproducible and consistent reasoning mode.
We discussed additional analysis about compute-performance trade-off in Section~\ref{sec:additional_analysis} and Table~\ref{tab:cost_analysis}.

\textbf{Relationship between transferability and Maj@K.}
We next study how reasoning transferability evolves over training by truncating a generator's reasoning trace into a prefix and continually generating a separate, frozen receiver model to complete it.
Specifically, we retain the first $t$ fraction of tokens with a fixed truncation ratio $t = 0.7$, and measure transferability as the receiver's verifiable final-answer accuracy under this cross-model continuation protocol.
Across training steps, \sname exhibits consistently stronger transferability growth than RLVR, with the gap generally widening as training progresses.
Notably, RLVR shows a clear drop in transferability at later training stages even when average accuracy remains stable, and this drop coincides with a degradation in Maj@64.
This coupling suggests that RLVR can become less consistent over training: it may preserve average correctness while increasingly relying on fragile or idiosyncratic traces that do not transfer well and thus fail to support self-consistency.
In contrast, \sname improves transferability together with Maj@64, implying that the transfer-based reward promotes reasoning traces that are both reusable under cross-model continuation and more reliably reproducible.
Overall, these dynamics provide direct evidence that the proposed transfer reward is effectively optimized during training, shaping the generator toward robust, ``transferable'' reasoning prefixes that remain useful beyond the original model's continuation.
Furthermore, additional results for performance in various truncation ratios, including $\tau\in\{0.3,0.5,0.7,0.9\}$, are illustrated in Appendix~\ref{app:transferability}.

\subsection{Component ablations}
In this section, we analyze the contribution of key design choices in \sname.
Specifically, we ablate two factors: (i) the reward composition (answer reward vs.\ transfer reward) and (ii) the receiver model.

\textbf{Effect of the reward ratio.}
To study the contribution of the transfer reward relative to the answer reward, we vary the weights $(a,t)$ in Eq.~\ref{eq:rlvr_reward} Eq.~\ref{eq:rltr_reward}, where $a$ scales the final-answer correctness reward and $t$ scales the transferability reward.
We evaluate the base model, RLVR $(a{=}1.0,\, t{=}0.0)$, and several \sname variants with different answer--transfer reward ratios: $(a{=}1.0,\, t{=}1.0)$, $(a{=}0.1,\, t{=}1.0)$, and $(a{=}1.0,\, t{=}0.1)$.

\begin{table}[t] 
    \vspace{0.03in}
    \caption{\textbf{Effect of reward ratio.} Average accuracy (\%) and Maj@K(\%) on MATH-500~\citep{hendrycks2021math} for \sname (Ours) by varying weight of transfer reward. Acc. denotes average accuracy. The bold indicates the best result within the group.} 
    \label{tab:reward_ratio}
    \vspace{0.05in}
    \small
    \centering
        \begin{tabular}{cc ccc c}
        \toprule
        &&\multicolumn{4}{c}{MATH-500}  \\
        \cmidrule(lr){3-6}
        a&t& Acc. & Maj@4& Maj@16& Maj@64 \\
        \midrule
        
        1.0&0.0&76.8& 78.2& 80.2& 82.2\\ 
        1.0&0.1&75.8& 77.4& 80.6& 81.2\\ 
        1.0&1.0&\textbf{77.0}& \textbf{81.0}& 83.2& 82.0\\ 
        0.1&1.0&\textbf{77.0}& 79.0& \textbf{83.8}& \textbf{84.2}\\ 
        \bottomrule
        \end{tabular}
    \vspace{-.07in}
\end{table}

As shown in Table~\ref{tab:reward_ratio}, increasing the transfer reward weight consistently improves performance in high-K regimes.
Models with larger $t$ achieve substantially higher Maj@K, while reducing the answer reward weight $a$ has a comparatively smaller effect.
In contrast, when the transfer reward is down-weighted $(t{=}0.1)$, the results closely resemble RLVR and exhibit limited scaling with K.
Assigning a larger transfer weight yields much stronger improvements: $(a{=}0.1,t{=}1.0)$ increases Maj@64 from 82.2 to 84.2.
Overall, these results suggest that emphasizing the transfer reward is important for improving reliability and scaling under large-K aggregation, beyond optimizing final-answer correctness alone.
This is consistent with our objective: encouraging transferable (i.e., reusable) reasoning prefixes makes independently sampled solutions more likely to converge to the same correct outcome, thereby improving consensus-driven metrics such as Maj@K in high-K regimes.
We further discuss a deeper analysis of multi-sampling performance in Appendix~\ref{app:ratio_pass}.

\textbf{Effect of receiver model.} 
The receiver model is used to continue truncated reasoning traces and thereby determines the transfer reward signal; thus, it can substantially affect training.
To isolate this factor, we vary the receiver while keeping the generator initialization and training data fixed.
Specifically, we compare three receivers (i.e., Llama3.2-3B-Instruct (Llama-3B)~\citep{grattafiori2024llama3}, Qwen2.5-1.5B-Instruct (Qwen-1.5B)~\citep{Yang2024Qwen25TR}, and Qwen2.5-3B-Instruct (Qwen-3B)~\citep{Yang2024Qwen25TR})and train \sname with the same generator and dataset.

\begin{table}[t] 
    \caption{\textbf{Receiver model ablation.} Average accuracy (\%) and Maj@K(\%) on MATH-500~\citep{hendrycks2021math} for \sname (Ours) by varying receiver model, including Base model, RLVR-trained model. Acc. denotes average accuracy.  The bold indicates the best result within the group.} 
    \label{tab:receiver_ablation}
    \vspace{0.05in}
    \small
    \centering
        \begin{tabular}{l c ccc }
        \toprule
        &\multicolumn{4}{c}{MATH-500}  \\
        \cmidrule(lr){3-5}
        Receiver& Acc. & Maj@4& Maj@16& Maj@64 \\
        \midrule
        Llama-3B&74.0& 78.4& 81.2& 82.0\\ 
        Qwen-1.5B&\textbf{77.6}& \textbf{79.6}& 82.4& 82.8\\ 
        Qwen-3B&77.0& 79.0& \textbf{83.8}& \textbf{84.2}\\ 
        \bottomrule
        \end{tabular}
\end{table}

As shown in Table~\ref{tab:receiver_ablation}, the receiver choice leads to noticeable differences, especially for majority voting at larger K.
For Maj@16 and Maj@64, Qwen-3B yields the best results (83.8 and 84.2), whereas Qwen-1.5B achieves the best Maj@4 (79.6).
This pattern suggests that a higher-capacity receiver can provide a more informative transfer signal that improves high-k consistency under majority voting.
Overall, using a stronger receiver tends to improve robustness under majority voting.
Notably, \sname still outperforms RLVR even with the smallest receiver (Qwen-1.5B).
These results indicate that the benefits of the transfer reward are not dependent on a large receiver, although stronger receivers further improve high-K consistency.
We further discuss a deeper analysis of multi-sampling performance in Appendix~\ref{app:receiver_pass}.

\subsection{Additional Analysis}\label{sec:additional_analysis}

\begin{table}[t]
    \caption{\textbf{Computational cost.} Comparison between RLVR and \sname with EFLOPs (i.e., $10^{18}$ FLOPs) and Average accuracy (Acc.) on MATH-500 evaluation.}
    \label{tab:cost_analysis}
    \vspace{0.05in}
    \small
    \centering
    \begin{tabular}{l ccc}
    \toprule
    & \multicolumn{3}{c}{MATH-500}\\
    \cmidrule(lr){2-4}
    & Train Steps& EFLOPs ($\downarrow$) & Acc. ($\uparrow$) \\
    \midrule

    RLVR& \phantom{0}40& \textbf{37.10} & 74.5 \\
    \textbf{\sname (Ours)} & \phantom{0}40& 39.76 & {76.2} \\
    RLVR& 100& 92.75 & {76.2} \\
    \textbf{\sname (Ours)} & \phantom{0}60& 59.64 & \textbf{76.8} \\
    \bottomrule
    \end{tabular}
    \vspace{-0.07in}
\end{table}
\textbf{Computational cost analysis.}
One possible concern of \sname is the extra computation introduced by the receiver (i.e., an additional model) during training. 
To this end, we demonstrate that, even with the extra computation, \sname still (i) achieves the same performance with less computation, and (ii) outperforms RLVR as the computational budget increases.
Here, we use FLOPs as the measure of computational cost, with the detailed computation procedure provided in Appendix ~\ref{app:efficiency}. 
As shown in Table~\ref{tab:cost_analysis}, \sname increases the training FLOPs by approximately 7\% under the same number of training steps, while achieving substantially better performance.
More importantly, we observed that RLVR requires 2.3$\times$ higher computational cost to reach the same performance level as \sname (39.76$\rightarrow$92.75 EFLOPs), highlighting both the sample efficiency and computational efficiency of \sname.
Lastly, it is worth noting that the accuracy should be regarded as the primary metric: \sname not only outperforms the best accuracy achieved by RLVR (at 92.75 EFLOPs), but does so with fewer training steps and lower computational cost.

\begin{table}[t]
    \vspace{0.03in}
    \caption{\textbf{\sname generalize beyond mathematical domain.} Accuracy (\%) and Maj@K (\%) on GPQA~\citep{rein2024gpqa}  for the base model and RL baselines. Acc. denotes average accuracy. The bold indicates the best result.}
    \label{tab:ood_gpqa}
    \vspace{0.05in}
    \small
    \centering
    \begin{tabular}{l cccc}
    \toprule
    & \multicolumn{4}{c}{GPQA} \\
    \cmidrule(lr){2-5}
    & Acc. & Maj@4 & Maj@8 & Maj@16 \\
    \midrule
    Base   & 32.4 & 35.3 & 35.0 & 35.2 \\
    RLVR   & 33.0 & 35.2 & 36.8 & 37.0 \\
    \rowcolor{RowHighlight}\textbf{\sname (Ours)} & \textbf{34.8} & \textbf{37.3} & \textbf{37.5} & \textbf{37.7} \\
    \bottomrule
    \end{tabular}
    \vspace{-0.09in}
\end{table}

\begin{table*}[t] 
    \caption{\textbf{\sname retains diversity better in notion of Pass@k.} Average accuracy (\%) and Pass@K (\%) for \sname (Ours) and other baselines, including Base model, and RLVR trained model. We consider two math reasoning benchmarks, MATH-500~\citep{hendrycks2021math}, and AMC23~\citep{maa23amc}. Acc. denotes average accuracy.  The bold indicates the best result within the group.} 
    \label{tab:pass}
    \vspace{0.05in}
    \small
    \centering
        \begin{tabular}{l cccc cccc}
        \toprule
        & \multicolumn{4}{c}{MATH-500} & \multicolumn{4}{c}{AMC23} \\
        \cmidrule(lr){2-5}\cmidrule(lr){6-9}

        Metrics&Acc.& Pass@4& Pass@16& Pass@64&Acc.& Pass@4& Pass@16& Pass@64\\
        \midrule
        Base Model& 71.0 & 86.2 & \textbf{92.4} & \textbf{95.2}& 46.2& 70.0 & 82.5 & \textbf{95.0}\\ 
        RLVR& 76.2 & 85.2 & 88.4 & 92.2& 52.8& 72.5 & 77.5 & 92.5\\ 
        \rowcolor{RowHighlight}\textbf{\sname (Ours)}& \textbf{77.0} & \textbf{88.2} & 92.2 & 95.0& \textbf{53.5}& \textbf{75.0} & \textbf{85.0} & \textbf{95.0}\\
        \bottomrule
        \end{tabular}
    \vspace{-.05in}
\end{table*}

\textbf{Performance on an out-of-domain reasoning task.} 
We further analyze to verify that \sname also enhances performance in out-of-domain, GPQA~\citep{rein2024gpqa}, whereas GPQA is a science question-answering benchmark that primarily tests logical and scientific reasoning beyond pure mathematics.
As shown in Table~\ref{tab:ood_gpqa},  \sname also yields consistent improvements in average accuracy and Maj@K. For example, average accuracy increased from 32.4 to 34.8 and Maj@16 from 35.2 to 37.7.
This suggests that the transfer reward is largely domain-agnostic, generalizing beyond mathematics to scientific reasoning.

\begin{figure}[t]
    \vspace{0.07in}
    \includegraphics[width=\columnwidth]{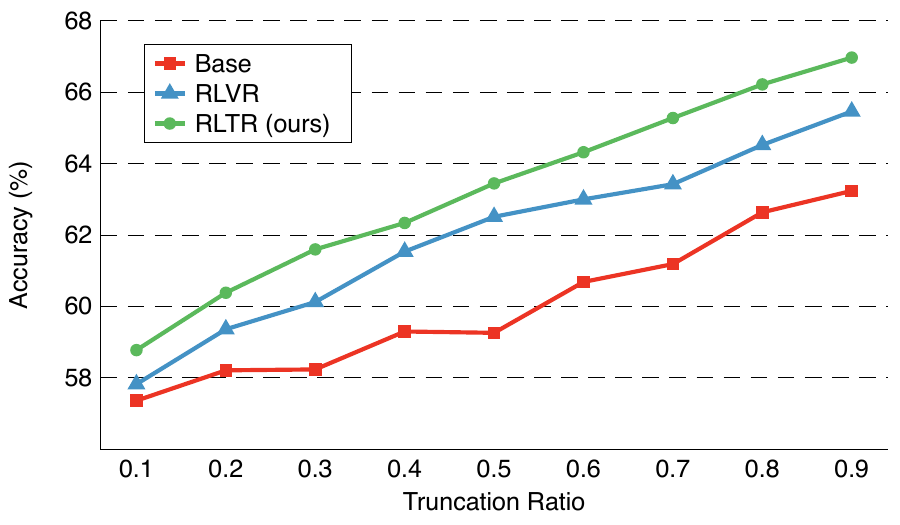}
    \vspace{-0.17in}
    \caption{
        \textbf{\sname improves transferability consistently.}
        Transferability (\%) comparison among base model, RLVR, and \sname.
        The truncation ratio is the fraction of tokens remained from the begging of a reasoning trace (higher means longer prefixes continuation).
        \sname consistently achieves higher transferability than baselines.
        } \label{figure:ratio_accuracy}
    \vspace{-0.07in}
\end{figure}

\textbf{Accuracy under reasoning truncation.} 
We evaluate reasoning robustness by measuring the transfer accuracy (i.e., transferability from the generator to Qwen2.5-3B-Instruct) of traces truncated prefix at various ratios on MATH-500.
As shown in Figure~\ref{figure:ratio_accuracy}, \sname consistently outperforms baselines (e.g., Base model and RLVR) across the entire spectrum $\tau$ from $0.1$ to $0.9$.
This demonstrates that \sname learns reasoning remains solvable under arbitrary interruptions, rather than exploiting specific truncation boundaries.
Furthermore, it is worth noting that \sname maintains a larger advantage even at small truncation ratios (e.g., $\tau\in[0.1,0.3]$), suggesting that it produces more informative prefixes from the very early stages of reasoning that provide sufficient intermediate structure for the receiver to complete the solution.

\textbf{Results on different LLM.} 
We also demonstrate the effectiveness of \sname on another LLM family. To this end, we train \sname from Llama3.1-8B-Instruct, where we use the receiver model as Llama3.2-3B-Instruct on the MATH-train dataset.
We report the average accuracy and majority voting accuracy (Maj@K) on the MATH-500 dataset.

\begin{table}[t] 
    \caption{\textbf{\sname is adaptable on other LLM architectures.} Average accuracy (\%) and Maj@K(\%) on MATH-500~\citep{hendrycks2021math} for \sname (Ours)  Llama-3.1-8B-Instruct. We compared the \sname with several baselines including Base model, RLVR trained model. 
    The bold indicates the best result within the group.} 
    \label{tab:ablation_llama}
    \vspace{0.05in}
    \small
    \centering
        \begin{tabular}{l cccc}
        \toprule

        Method& Acc. & Maj@4& Maj@16 & Maj@64\\
        \midrule
        Base Model & 24.8&34.4&51.2&60.8\\ 
        RLVR& \textbf{50.0} & \textbf{55.2} & 58.6 & 59.8\\ 
        \rowcolor{RowHighlight}\textbf{\sname (Ours)} & 45.0&52.6&\textbf{59.2}&\textbf{63.4}\\ 
        \bottomrule
        \end{tabular}
    \vspace{-.07in}
\end{table}
As shown in Table~\ref{tab:ablation_llama}, \sname outperforms the baselines at high-K Maj@K even when trained with a different LLM family.
For example, \sname achieves Maj@16 of 59.2 and Maj@64 of 63.4, substantially higher than both the base model and RLVR-trained model.
In contrast, RLVR primarily improves single-sample accuracy; however, these gains do not translate into sampling consistency, as performance improvements saturate under high-$K$ majority voting and even underperform the base model at Maj@64.
These results are consistent with the main findings in Section~\ref{sec:experiment_main}, highlighting the importance of robust reasoning, which is also demonstrated across different LLM families.

\textbf{Pass@K Evaluation.} We have discussed the consistency and robustness of the reasoning process in prior sections. Now, we analyze the diversity aspect of the training. We quantify the meaningful diversity of the reasoning using the Pass@K metric. Pass@K measures whether at least one of the K sampled solutions contains the correct final answer.

As shown in Table~\ref{tab:pass}, RLVR improves single-sample accuracy on MATH-500, but its high-K diversity degrades compared to the Base model (e.g., Pass@64: 95.2$\to$92.2).
In contrast, \sname preserves strong high-K Pass@K while maintaining competitive single-sample average accuracy (Acc.) on MATH-500, \sname achieves comparable Pass@K performance, while outperforming other baselines in single-sample average accuracy.
A similar pattern holds on AMC23, where RLVR again reduces high-K Pass@K, whereas \sname consistently improves or matches the Base model across all K (e.g., Pass@64: 95.0).
These results suggest that conventional RLVR tends to concentrate probability mass on a narrower set of solution patterns, improving average single-sample accuracy but harming diversity, consistent with observations in~\citet{yue2025does}.
By contrast, incorporating transfer-based signals enables \sname to improve robustness while preserving meaningful diversity in sampled solutions.

\section{Conclusion}
\label{sec:con}
We introduced \mname, a transfer-augmented RLVR framework that optimizes \emph{reasoning transferability}, the ability of a truncated reasoning prefix to be completed by another model to a verifiably correct answer.
By leveraging a transfer reward, \sname encourages reasoning traces that remain reusable under truncation and cross-model continuation.
Across multiple reasoning benchmarks, \sname improves average accuracy and sampling-based consistency (e.g., majority voting) indicating enhanced robustness of learned reasoning processes.
Moreover, \sname significantly improves sample efficiency which reduce training steps to achieve comparable performance with baseline.
We further shows robust generalization across base models and out-of-distribution settings.
Overall, our results suggest that optimizing transferability leads to more robust LLM reasoning.

\section*{Impact Statement}
This work aims to advance robust LLM reasoning by introducing transferability as a training signal.
We show that \sname can improve not only accuracy but also reasoning consistency, while significantly enhancing sample efficiency and enabling faster learning.

Today, training and sample efficiency is an important goal to both research and industry, because of the increasing computation cost and concerns about sustainable AI development.
By improving the consistency and sample efficiency, our approach has a potential to benefit to the communities.

Furthermore, in the application side, domains that require consistency and heavily annotated training samples (e.g., Vision Language Action (VLA) model or Computer Use Agent (CUA) ) can easily adopt our method by augmenting the existing RLVR frameworks. 
Consequently, by enabling reliable reaosning in such complex applications, our method may contribute to the societal/technological innovation.

\bibliography{ref}
\bibliographystyle{icml2026}
\clearpage
\appendix
\onecolumn
\section{Experimental Details}
In this section, we describe the experimental details of Section \ref{sec:experiments}, including \sname and the baselines.
\subsection{Dataset Details}\label{app:dataset_detail}
Here, we describe the dataset we used in training and evaluation.
\begin{itemize}[itemsep=0.05in,leftmargin=0.2in,topsep=0.02in]
    \item{\textbf{GSM8K.}} 
    The GSM8K dataset~\citep{cobbe2021gsm8k} consists of 7,473 train samples and 1,319 test samples of moderate-level of math problems. This is diverse grade school math word problems. We used test set in our experiments which consists of problem and answer pairs.
    \item{\textbf{MATH.}} 
    The MATH dataset~\citep{hendrycks2021math} consists of 7,500 train samples and 5,000 test samples. This dataset consists of diverse range of difficulty level. We utilized 3K subset from MATH train-set which has high difficulty (>=3). Also, for the evaluation we used MATH-500 which is subset of test-set~\citep{lightman2023lets}.
    \item{\textbf{AMC23.}}
    The AMC23 dataset~\citep{maa23amc} consists of 40 competition-level hard problems, provided as linguistic problem and answer pairs. We used this dataset for evaluation in our experiments.
    \item{\textbf{AIME2024.}}
    The AIME2024 dataset~\citep{maa24aime} consists of 30 competition-level hard problems, provided as linguistic problem and answer pairs. We used this dataset for evaluation in our experiments. This is the most difficult benchmark among our evaluation sets.
    \item{\textbf{GPQA.}}
    The GPQA dataset~\citep{rein2024gpqa} consists of 448 multiple-choice question and answer pairs. GPQA is a science question-answering benchmark that primarily tests logical and scientific reasoning beyond pure mathematics. We used the GPQA dataset for out-of-distribution evaluation in our experiments.
\end{itemize}

\subsection{Model Details}\label{app:model_detail}
Here, we describe the model and their source we used in training and evaluation. We utlized checkpoints provided in huggingface.
\begin{itemize}[itemsep=0.05in,leftmargin=0.2in,topsep=0.02in]
    \item{\textbf{Qwen2.5-7B-Instruct.}} (\citealp{Yang2024Qwen25TR};https://huggingface.co/Qwen/Qwen2.5-7B-Instruct)
    \item{\textbf{Qwen2.5-3B-Instruct.}} (\citealp{Yang2024Qwen25TR};https://huggingface.co/Qwen/Qwen2.5-3B-Instruct)
    \item{\textbf{Qwen2.5-1.5B-Instruct.}} (\citealp{Yang2024Qwen25TR}; https://huggingface.co/Qwen/Qwen2.5-1.5B-Instruct) 
    \item{\textbf{Llama3.1-8B-Instruct.}}(\citealp{grattafiori2024llama3}; https://huggingface.co/meta-llama/Llama-3.1-8B-Instruct)
    \item{\textbf{Llama3.2-3B-Instruct.}} (\citealp{grattafiori2024llama3}; https://huggingface.co/meta-llama/Llama-3.2-3B-Instruct)
\end{itemize}

\subsection{Training Details}\label{app:training_detail}
\textbf{Training framework.} For the training framework, we utilize VERL~\citep{sheng2024hybridflow} for both \sname and RLVR.

\begin{itemize}[itemsep=0.05in,leftmargin=0.2in,topsep=0.02in]
\item VERL (https://github.com/volcengine/verl)
\end{itemize}

\textbf{Training hyperparameters.} We describe our training hyperparameters in Table~\ref{tab:hyperparameters_rl}, which we used for training \sname and RLVR.
\begin{table}[t]
  \centering
  \caption{Hyperparameters for \sname and RLVR training. ($^*$ denotes the hyperparameter only used for \sname.)}
  \label{tab:hyperparameters_rl}
  \vspace{0.05in}
  \begin{tabular}{l | l}
    \toprule
    \textbf{Hyperparmeter} & \textbf{Value} \\
    \midrule
    Optimizer & Adam\\
    Algorithm & GRPO\\
    Training steps & 150\\ 
    Learning rate & 1e-6\\
    Batch size & 1,024\\
    Generation per sample & 8\\
    kl\_coef & 0.001\\
    Generator temperature & 1.0 \\
    Receiver temperature$^*$ & 1.0 \\
    Max new generation length & 8,192\\
    
    \bottomrule
    \end{tabular}
  \vspace{-0.15in}
\end{table}

\textbf{Compute Resource.}\label{appendix:compute_resource}
For the main development, we mainly use eight H200 141 GB GPUs. 

\subsection{Evaluation Details}\label{app:evalution_detail}
\textbf{Evaluation Framework. } For the evaluation framework, we utilize Math Verify and SimpleRL~\citep{zeng2025simplerl}.

\begin{itemize}[itemsep=0.05in,leftmargin=0.2in,topsep=0.02in]
\item{\textbf{SimpleRL.}} 
(https://github.com/hkust-nlp/simpleRL-reason)
\item{\textbf{Math Verify.}}
(https://github.com/huggingface/Math-Verify)
\end{itemize}

\section{Additional Results}

\subsection{Rigorous Analysis of Computation Efficiency of \sname}\label{app:efficiency}
In this section, we provide analysis of the computational cost of \sname compared to the standard RLVR baseline. Ultimately, we argue that while \sname has overhead in per step computational cost, this is outweighted by its training step efficiency, leading to a significant advantage in terms of total FLOPs to reach comparable performance.

To rigorously assess the training cost, we estimate the Floating Point Operations (FLOPs) per token following standard scaling laws~\citep{shao2024deepseekmath}. Specifically, we followed the \citet{xiaoju2023calflops} to calculate the FLOPs precisely.

\begin{table}[h]
\centering
\caption{Model architectures for Qwen2.5-7B-Instruct~\citep{Yang2024Qwen25TR} and Qwen2.5-3B-Instruct~\citep{Yang2024Qwen25TR}.}
\vspace{0.05in}
\label{tab:model_specs}
\begin{tabular}{lcc}
\toprule
\textbf{Hyperparameter} & \textbf{Qwen2.5-7B-Instruct} & \textbf{Qwen2.5-3B-Instruct} \\
\midrule
Parameters & $\sim$7.6B & $\sim$3.1B \\
Hidden Size ($d_{model}$) & 3,584 & 2,048 \\
Layers ($L$) & 28 & 36 \\
Attention Heads & 28 & 16 \\
Vocab Size ($V$) & 152,064 & 151,936 \\
\bottomrule
\end{tabular}
\vspace{-0.05in}
\end{table}
\begin{table}[h]
\centering
\caption{FLOPs breakdown per training step. RLTR adds a receiver rollout phase compared to RLVR.}
\label{tab:flops}
\vspace{0.05in}
\begin{tabular}{lrr}
\toprule
Component & RLVR & RLTR (Ours) \\
& \small{(PFLOPs/step)} & \small{(PFLOPs/step)} \\
\midrule
Policy Rollout (Generation) & 141.65 & 141.65 \\
Policy Forward (Logprobs) & 157.17 & 157.17 \\
Reference Forward (KL) & 157.17 & 157.17 \\
Policy Training (Fwd+Bwd) & 471.50 & 471.50 \\
\quad \textbf{Receiver Rollout (Transfer)} & \textbf{—} & \textbf{66.49} \\
\midrule
\textbf{Total FLOPs per Step} & \textbf{927.48} & \textbf{993.96} \\
\bottomrule
\end{tabular}
\vspace{-0.015in}
\end{table}

First, we describe the model architectures of Qwen2.5-7B-Instruct and Qwen2.5-3B-Instruct, which were mainly used in our experiments, in Table~\ref{tab:model_specs}. 
For these calculations we set the hyperparameters as described in Table~\ref{tab:hyperparameters_rl}. 
Since we adopts GRPO, there are four primary computational phases, policy rollout, policy forward (Logprobs), reference forward (KL), and policy training (forward and backward passes). Specifically, we calculated the FLOPs for forward, backward, and autoregressive generation operations using the formulas provided by \citet{xiaoju2023calflops}.

Also, in the case \sname, an additional computation needs for the transfer rollout using the smaller receiver model (i.e., Qwen2.5-3B-Instruct).
Specifically, as shown in Table~\ref{tab:flops}, the total FLOPs per step is 927.48 PFLOPs for RLVR and 993.96 PFLOPs for \sname , which results in approximately 7.2\% overhead.
However, this per-step overhead is negligible compared to the gains in sample efficiency. 
As demonstrated in Figure~\ref{fig:avg_progress} of the main text, \sname converges to the same accuracy level approximately \textbf{60\% reduced} than RLVR. 
Consequently, the \textbf{total computational cost} to train \sname is \textbf{reduced by roughly 57\%} compared to the baseline, making it a highly efficient framework for robust reasoning training.

\begin{figure*}[t]
  \centering
    \begin{subfigure}[t]{0.49\textwidth}
    \centering
    \includegraphics[width=\linewidth]{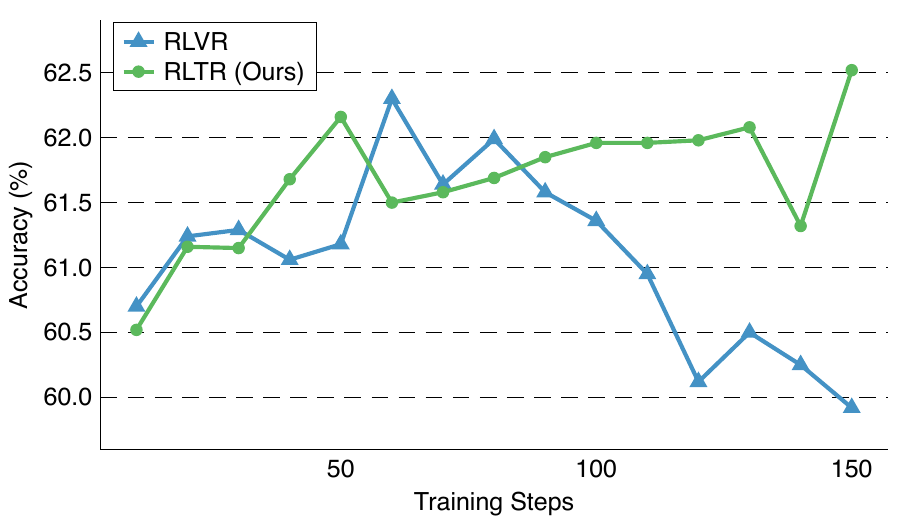}
    \caption{\textbf{Truncation Ratio ($\tau$): 0.3}}
    \label{fig:app_transfer_3}
  \end{subfigure}\hfill
  \begin{subfigure}[t]{0.49\textwidth}
    \centering
    \includegraphics[width=\linewidth]{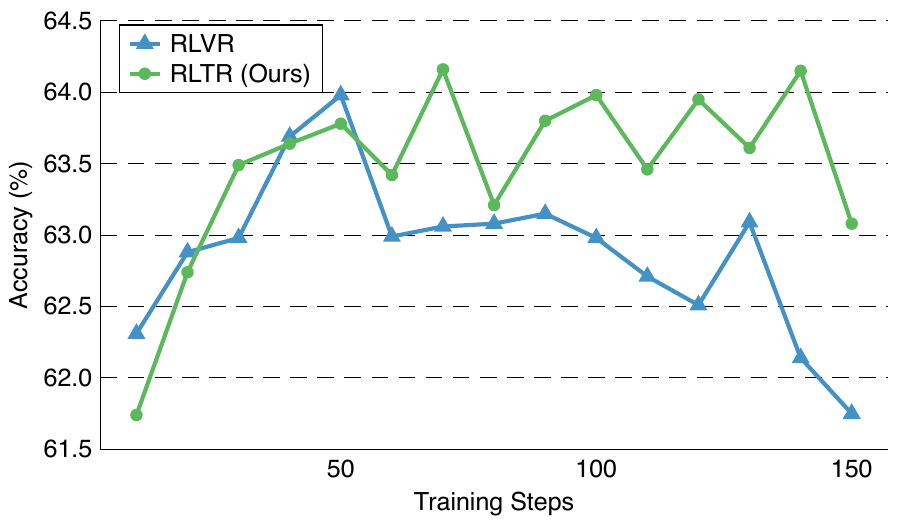}
    \caption{\textbf{Truncation Ratio ($\tau$): 0.5}}
    \label{fig:app_transfer_5}
  \end{subfigure}\hfill
  \begin{subfigure}[t]{0.49\textwidth}
    \centering
    \includegraphics[width=\linewidth]{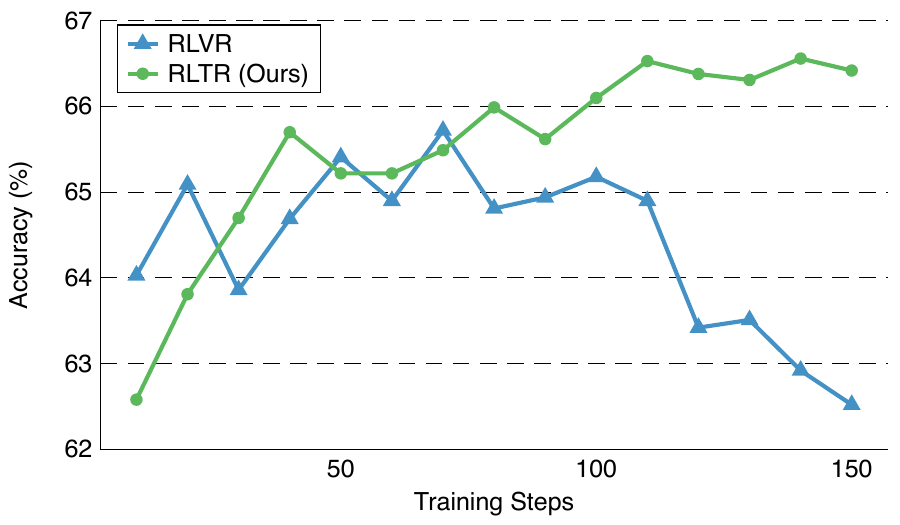}
    \caption{\textbf{Truncation Ratio ($\tau$): 0.7}}
    \label{fig:app_transfer_7}
  \end{subfigure}
  \begin{subfigure}[t]{0.49\textwidth}
    \centering
    \includegraphics[width=\linewidth]{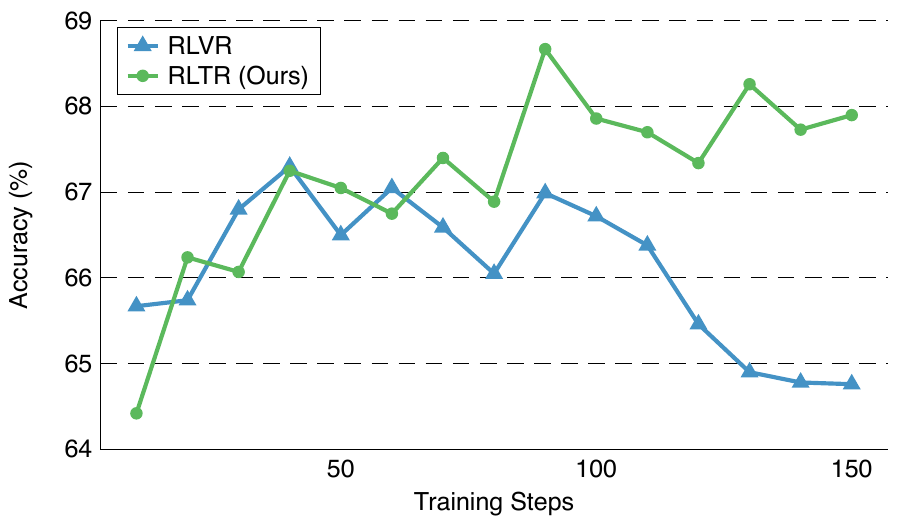}
    \caption{\textbf{Truncation Ratio ($\tau$): 0.9}}
    \label{fig:app_transfer_9}
  \end{subfigure}

  \caption{
  \textbf{Detailed Analysis of transferability on various truncation ratios $\tau\in\{0.3,0.5,0.7,0.9\}$.}
  }  
  \label{fig:app_transfer}
\end{figure*}
\subsection{Additional Analysis for Transferability}\label{app:transferability}
We extended our analysis of training dynamics of transferability, which is described in Figure~\ref{fig:transfer_progress}.
We reported the transferability dynamics and set the truncation ratio $\tau=0.7$ in the main part.
Here, we vary the truncation ratio from $\tau\in\{0.3,0.5,0.7,0.9\}$ while keeping the same settings.
As shown in Figure~\ref{fig:app_transfer}, the transferability gap between RLVR and \sname is consistently getting larger through all truncation ratios.
This suggests that \sname learns a robust reasoning process which is unspecified to a certain truncation ratio boundary.

\subsection{Additional Analysis on Reward Ratio}\label{app:ratio_pass}
\begin{table}[t] 
    \vspace{-0.1in}
    \caption{\textbf{Effect of Reward Ratio.} Accuracy (\%) for \sname (Ours) by varying weight of transfer reward, including Base model, Baset-RL trained model. We consider math reasoning benchmarks, MATH-500~\citep{hendrycks2021math}. The bold indicates the best result within the group.} 
    \label{tab:reward_ratio_pass}
    \vspace{0.05in}
    \small
    \centering
        \begin{tabular}{cc cccc}
        \toprule

        &&\multirow{2.5}{*}{Acc.} & \multicolumn{3}{c}{Pass@K} \\
        \cmidrule(lr){4-6}
        a&t& & 4& 16& 64\\
        \midrule
        
        1.0&0.0&76.8& 85.2& 89.8& 92.2\\ 
        1.0&0.1&75.8& 84.4& 89.0& 92.4\\ 
        1.0&1.0&\textbf{77.0}& 85.2& 89.4& 92.6\\ 
        0.1&1.0&\textbf{77.0}& \textbf{88.2}& \textbf{92.2}& \textbf{95.0}\\ 
        \bottomrule
        \end{tabular}
    \vspace{-.07in}
\end{table}

We extended our analysis on the reward ratio model, which is discussed in Table~\ref{tab:reward_ratio}. Specifically, we analyze diversity in the notion of Pass@K. As shown in Table~\ref{tab:reward_ratio_pass}, assigning a larger transfer weight yields much stronger improvements: $(a{=}0.1,t{=}1.0)$ increases Pass@64 from 92.2 to 95.0. Overall, larger transfer weight improves diversity, indicating that our transfer reward approach can also help retain diversity degrade problem of RLVR, which was suggested in \citet{yue2025does}.

\subsection{Additional Analysis on Reveiver Model}\label{app:receiver_pass}

\begin{table}[t] 
    \caption{\textbf{Receiver Model Ablation.} Accuracy (\%) for \sname (Ours) by varying receiver model, including Base model, RLVR trained model. We consider a math reasoning benchmark, MATH-500~\citep{hendrycks2021math}. The bold indicates the best result within the group.} 
    \label{tab:receiver_ablation_pass}
    \vspace{0.05in}
    \small
    \centering
        \begin{tabular}{l c ccc}
        \toprule
        &\multirow{2.5}{*}{Acc.} & \multicolumn{3}{c}{Pass@K} \\
        \cmidrule(lr){3-5}
        Receiver& & 4& 16& 64\\
        \midrule
        Llama-3B&74.0& 83.8& 91.8& \textbf{95.2}\\ 
        Qwen-1.5B&\textbf{77.6}& 87.6& 91.0& 95.0\\ 
        Qwen-3B&77.0& \textbf{88.2}& \textbf{92.2}& 95.0\\ 
        \bottomrule
        \end{tabular}
    \vspace{-.07in}
\end{table}

We extended our analysis on the receiver model, which is discussed in Table~\ref{tab:receiver_ablation}. Specifically, we analyze diversity in the notion of Pass@K. As shown in Table~\ref{tab:receiver_ablation_pass}, all receivers achieve similar performance at large $K$ (Pass@64: 95.2 vs.\ 95.0 vs.\ 95.0), indicating comparable diversity in producing at least one correct sample.
However, average accuracy varies more across receivers (74.0/77.6/77.0), suggesting that receivers can influence single-sample correctness while preserving diversity.
Overall, the diversity captured by Pass@K is largely preserved.

\section{Prompt Examples}
In this section, we describe the prompt template used for training and evaluation in our experiments. As shown in Table~\ref{tab:appendix_prompt_template}, we instruct the model to reason before generating the final answer. Moreover, as shown in Table~\ref{tab:appendix_prompt_template_gpqa}, for GPQA, which provides four choices per problem, we included these options in the prompt in a random order. For training and evaluation, we place these prompt in user turn of the model-specific chat templates.
\begin{table}[h]
    \caption{The user templates for MATH~\citep{hendrycks2021math}, GSM8K~\citep{cobbe2021gsm8k}, AMC23~\citep{maa23amc}, and \citep{maa24aime}. We used it for training and evaluation.}
    \vspace{0.05in}
    \label{tab:appendix_prompt_template}
    \small
    \centering
    \begin{tabular}{p{14.5cm}}
        \toprule
        \textbf{User template for training and evaluation.} \\
    \end{tabular}
    
    \footnotesize
    \begin{tabular}{p{14.5cm}}
        \midrule
        \{Problem\}\\
        You FIRST think about the reasoning process as an internal monologue and then provide the final answer.\\
        The final answer MUST BE put in \textbackslash{}boxed\{\}.\\
        \bottomrule
    \end{tabular}
\end{table}

\begin{table}[h]
    \caption{The user template for GPQA~\citep{rein2024gpqa}.}
    \vspace{0.05in}
    \label{tab:appendix_prompt_template_gpqa}
    \small
    \centering
    \begin{tabular}{p{14.5cm}}
        \toprule
        \textbf{User template for evaluating the GPQA.} \\
    \end{tabular}
    
    \footnotesize
    \begin{tabular}{p{14.5cm}}
        \midrule
        \{Problem\}\\
        You FIRST think about the reasoning process as an internal monologue and then provide the final answer.\\
        This is a multiple-choice question with options A, B, C, D.\\
        Your final answer MUST be ONLY the letter of the correct option (A, B, C, or D).\\
        Put your final answer in \textbackslash{}boxed\{\}, for example: \textbackslash{}boxed\{A\} or \textbackslash{}boxed\{B\}.\\
        \\
        Options:\\
        A. \{Answer1\}\\
        B. \{Answer2\}\\
        C. \{Answer3\}\\
        D. \{Answer4\}\\
        \bottomrule
    \end{tabular}
\end{table}

\section{Future Work}
\textbf{Transfer as a verifier.} An interesting next step is to leverage reasoning transferability beyond training as a test-time scaling mechanism.
Since our results suggest that optimizing for transferability yields more consistent and reusable reasoning traces, a natural extension is to incorporate transferability into the verification stage itself.
This would enable a verification module that explicitly prefers solutions whose partial reasoning is robust under cross-model continuation.


\section{Generation Examples}
We compare generation exmples from RLVR-trained model and a \sname-trained model.
As shown in Tables~\ref{tab:app_gen_rltr} and \ref{tab:app_gen_rlvr}, both models solve same problem and the final answer is correct.
However, the underlying reasoning differs substantially.
The RLVR-trained model contains a clear inconsistency in an intermediate reasoning trajectory (highlight in red).
In contrast, \sname produces a coherent and logically consistent trajectory.
This example supports our hypothesis that optimizing only final-answer correctness in RLVR can fail to penalize brittle reasoning  traces, while \sname successfully incentivize to capture the flawless in the reasoning process
\begin{table}[h]
    \caption{Generation Example of \sname-trained model on MATH-500~\citep{hendrycks2021math}.}\label{tab:app_gen_rltr}
    \vspace{0.05in}
    \small
    \centering
    \begin{tabular}{p{14.5cm}}
        \toprule
Define
\[p = \sum_{k = 1}^\infty \frac{1}{k^2} \quad \text{and} \quad q = \sum_{k = 1}^\infty \frac{1}{k^3}.\]Find a way to write
\[\sum_{j = 1}^\infty \sum_{k = 1}^\infty \frac{1}{(j + k)^3}\]in terms of $p$ and $q.$\\
\midrule
To find the value of the double sum \(\sum_{j = 1}^\infty \sum_{k = 1}^\infty \frac{1}{(j + k)^3}\) in terms of \(p\) and \(q\), we start by considering the inner sum. We can rewrite the double sum as follows:
\[
\sum_{j = 1}^\infty \sum_{k = 1}^\infty \frac{1}{(j + k)^3}.
\]
We can change the order of summation by considering the sum over all possible values of \(n = j + k\). For a fixed \(n\), \(j\) can range from 1 to \(n-1\) and \(k = n - j\). Therefore, the double sum can be rewritten as:
\[
\sum_{n = 2}^\infty \sum_{j = 1}^{n-1} \frac{1}{n^3}.
\]
The inner sum is simply \((n-1) \cdot \frac{1}{n^3}\), so we have:
\[
\sum_{n = 2}^\infty \frac{n-1}{n^3} = \sum_{n = 2}^\infty \left( \frac{n}{n^3} - \frac{1}{n^3} \right) = \sum_{n = 2}^\infty \left( \frac{1}{n^2} - \frac{1}{n^3} \right).
\]
We can split this into two separate sums:
\[
\sum_{n = 2}^\infty \frac{1}{n^2} - \sum_{n = 2}^\infty \frac{1}{n^3}.
\]
We can express these sums in terms of \(p\) and \(q\). Note that:
\[
\sum_{n = 2}^\infty \frac{1}{n^2} = \sum_{n = 1}^\infty \frac{1}{n^2} - \frac{1}{1^2} = p - 1,
\]
and
\[
\sum_{n = 2}^\infty \frac{1}{n^3} = \sum_{n = 1}^\infty \frac{1}{n^3} - \frac{1}{1^3} = q - 1.
\]
Therefore, the original sum becomes:
...
Thus, the value of the double sum is:
\[
\boxed{p - q}.
\]\\

        \bottomrule
    \end{tabular}
\end{table}

\begin{table}[h]
    \caption{Generation Example of RLVR-trained model on MATH-500~\citep{hendrycks2021math}. Highlighted errorneous part in red.}\label{tab:app_gen_rlvr}
    \vspace{0.05in}
    \small
    \centering
    \begin{tabular}{p{14.5cm}}
        \toprule

We need to evaluate the double sum \(\sum_{j=1}^\infty \sum_{k=1}^\infty \frac{1}{(j+k)^3}\) in terms of \(p\) and \(q\), where 
\[ p = \sum_{k=1}^\infty \frac{1}{k^2} \]
and 
\[ q = \sum_{k=1}^\infty \frac{1}{k^3}. \]

1. **Rewriting the Double Sum:** 
   \[
   \sum_{j=1}^\infty \sum_{k=1}^\infty \frac{1}{(j+k)^3}
   \]

2. **Index Transformation:**
   Change the order of summation by letting \( n = j + k \). Thus, for a fixed \( n \), \( j \) can range from 1 to \( n-1 \) (since \( j \) and \( k \) are both positive integers and \( k = n - j \)) and \( k \) can range from 1 to \( n-1 \). The sum becomes the sum over \( n \) and the corresponding values of \( j \) and \( k \):
   \[
   \sum_{j=1}^\infty \sum_{k=1}^\infty \frac{1}{(j+k)^3} = \sum_{n=2}^\infty \sum_{j=1}^{n-1} \frac{1}{n^3}.
   \]

3. **Inner Sum Evaluation:**
   The inner sum over \( j \) for a fixed \( n \) is:
   \[
   \sum_{j=1}^{n-1} \frac{1}{n^3} = \frac{1}{n^3} \sum_{j=1}^{n-1} 1 = \frac{1}{n^3} (n-1).
   \]

4. **Simplification:**
   Substituting the inner sum back into the outer sum, we get:
{\color{red}   \[
   \sum_{n=2}^\infty \sum_{j=1}^{n-1} \frac{1}{(j+k)^3} = \sum_{n=2}^\infty \frac{n-1}{n^3}.
   \]
}
5. **Further Simplification of the Sum:**
   Now, we need to simplify \(\sum_{n=2}^\infty \frac{n-1}{n^3}\):
   \[
   \sum_{n=2}^\infty \frac{n-1}{n^3} = \sum_{n=2}^\infty \left( \frac{n}{n^3} - \frac{1}{n^3} \right) = \sum_{n=2}^\infty \frac{1}{n^2} - \sum_{n=2}^\infty \frac{1}{n^3}.
   \]

6. **Separating the Series:**
   We split the series into two parts:
   \[
   \sum_{n=2}^\infty \frac{1}{n^2} = \left( \sum_{n=1}^\infty \frac{1}{n^2} \right) - 1 = p - 1.
   \]
   And for the second part:
   \[
   \sum_{n=2}^\infty \frac{1}{n^3} = \left( \sum_{n=1}^\infty \frac{1}{n^3} \right) - 1 = q - 1.
   \]

7. **Combining the Results:**
   Substituting these back, we get:
   \[
   \sum_{n=2}^\infty \frac{n-1}{n^3} = (p - 1) - (q - 1) = p - q.
   \]

Therefore, the expression for the sum \(\sum_{j=1}^\infty \sum_{k=1}^\infty \frac{1}{(j+k)^3}\) in terms of \(p\) and \(q\) is:
\[
\boxed{p - q}.
\]\\
        \bottomrule
    \end{tabular}
\end{table}

\end{document}